\documentclass[runningheads]{llncs}
\usepackage{amsfonts}
\usepackage{graphicx}
\usepackage{amsfonts}
\usepackage{graphicx}
\usepackage{epsfig}
\usepackage{graphicx}
\usepackage{amsmath}
\usepackage{amssymb}
\usepackage{mathtools}
\usepackage{resizegather}
\usepackage{marvosym}
\usepackage{dsfont}
\usepackage{enumitem}
\usepackage{caption}
\usepackage{multirow}
\RequirePackage[utf8]{inputenc}

\usepackage[colorlinks=true, allcolors=blue]{hyperref}

\usepackage[ruled, vlined]{algorithm2e}

\usepackage{lipsum}
\usepackage{array}
\usepackage{amsmath}
\DeclareMathOperator*{\argmax}{arg\,max}

\usepackage{enumitem}
 
\begin{document}

\title{ACT: Semi-supervised Domain-adaptive Medical Image Segmentation with Asymmetric Co-Training}
\titlerunning{ACT for SSDA Segmentation}

\author{Xiaofeng Liu\inst{1} \and Fangxu Xing\inst{1} \and Nadya Shusharina\inst{2} \and Ruth Lim\inst{1} \and C.-C. Jay Kuo\inst{3}  \and Georges El Fakhri\inst{1}  \and Jonghye Woo\inst{1}}

%index{Xiaofeng, Liu}
%index{Fangxu, Xing}
%index{Nadya Shusharina}
%index{Ruth, Lim}
%index{C.-C. Jay, Kuo}
%index{Georges, El Fakhri}
%index{Jonghye, Woo}

\institute{Gordon Center for Medical Imaging, Department of Radiology, Massachusetts General Hospital and Harvard Medical School, Boston, MA, 02114\and
Division of Radiation Biophysics, Department of radiation Oncology, Massachusetts General Hospital and Harvard Medical School, Boston, MA, 02114\and
Department of Electrical and Computer Engineering, University of Southern California, Los Angeles, CA, 90007}
 
\authorrunning{X. Liu et al.}

\maketitle              % typeset the header of the contribution
  
\begin{abstract}

Unsupervised domain adaptation (UDA) has been vastly explored to alleviate domain shifts between source and target domains, by applying a well-performed model in an unlabeled target domain via supervision of a labeled source domain. Recent literature, however, has indicated that the performance is still far from satisfactory in the presence of significant domain shifts. Nonetheless, delineating a few target samples is usually manageable and particularly worthwhile, due to the substantial performance gain. Inspired by this, we aim to develop semi-supervised domain adaptation (SSDA) for medical image segmentation, which is largely underexplored. We, thus, propose to exploit both labeled source and target domain data, in addition to unlabeled target data in a unified manner. Specifically, we present a novel asymmetric co-training (ACT) framework to integrate these subsets and avoid the domination of the source domain data. Following a divide-and-conquer strategy, we explicitly decouple the label supervisions in SSDA into two asymmetric sub-tasks, including semi-supervised learning (SSL) and UDA, and leverage different knowledge from two segmentors to take into account the distinction between the source and target label supervisions. The knowledge learned in the two modules is then adaptively integrated with ACT, by iteratively teaching each other, based on the confidence-aware pseudo-label. In addition, pseudo label noise is well-controlled with an exponential MixUp decay scheme for smooth propagation. Experiments on cross-modality brain tumor MRI segmentation tasks using the BraTS18 database showed, even with limited labeled target samples, ACT yielded marked improvements over UDA and state-of-the-art SSDA methods and approached an ``upper bound" of supervised joint training.

\end{abstract}

\section{Introduction} 

Accurate delineation of lesions or anatomical structures is a vital step for clinical diagnosis, intervention, and treatment planning \cite{tajbakhsh2020embracing}. While recently flourished deep learning methods excel at segmenting those structures, deep learning-based segmentors cannot generalize well in a heterogeneous domain, e.g., different clinical centers, scanner vendors, or imaging modalities \cite{liu2022deep,Liu_2021_ICCV,liuconstraining,che2019deep}. To alleviate this issue, unsupervised domain adaptation (UDA) has been actively developed, by applying a well-performed model in an unlabeled target domain via supervision of a labeled source domain \cite{chen2019synergistic,liu2021domain,liu2021generative,liu2022unsupervisedFrontiers}. Due to diverse target domains, however, the performance of UDA is far from satisfactory \cite{zou2020unsupervised,han2022deep,liu2022self}. Instead, labeling a small set of target domain data is usually more feasible \cite{van2020survey}. As such, semi-supervised domain adaptation (SSDA) has shown great potential as a solution to domain shifts, as it can utilize both labeled source and target data, in addition to unlabeled target data. To date, while several SSDA classification methods have been proposed~\cite{donahue2013semi,yao2015semi,saito2019semi,kim2020attract}, based on discriminative class boundaries, they cannot be directly applied to segmentation, since segmentation involves complex and dense pixel-wise predictions.

Recently, while a few works \cite{wang2020alleviating,chen2021semi,hoyer2021improving} have been proposed to extend SSDA for segmentation on natural images, to our knowledge, no SSDA for medical image segmentation has yet been explored. For example, a depth estimation for natural images is used as an auxiliary task as in \cite{hoyer2021improving}, but that approach cannot be applied to medical imaging data, e.g., MRI, as they do not have perspective depth maps. Wang et al.~\cite{wang2020alleviating} simply added supervision from labeled target samples to conventional adversarial UDA. Chen et al.~\cite{chen2021semi} averaged labeled source and target domain images at both region and sample levels to mitigate the domain gap. However, source domain supervision can easily dominate the training, when we directly combine the labeled source data with the target data \cite{saito2019semi}. In other words, the extra small amount of labeled target data has not been effectively utilized, because the volume of labeled source  data is much larger than labeled target data, and there is significant divergence across domains \cite{saito2019semi}.

To mitigate the aforementioned limitations, we propose a practical asymmetric co-training (ACT) framework to take each subset of data in SSDA in a unified and balanced manner. In order to prevent a segmentor, jointly trained by both domains, from being dominated by the source data only, we adopt a divide-and-conquer strategy to decouple the label supervisions for the two asymmetric segmentors, which share the same objective of carrying out a decent segmentation performance for the unlabeled data. By ``asymmetric," we mean that the two segmentors are assigned different roles to utilize the labeled data in either source or target domain, thereby providing a complementary view for the unlabeled data. That is, the first segmentor learns on the labeled source domain data and unlabeled target domain data as a conventional UDA task, while the other segmentor learns on the labeled and unlabeled target domain data as a semi-supervised learning (SSL) task. To integrate these two asymmetric branches, we extend the idea of co-training \cite{blum1998combining,balcan2005co,qiao2018deep}, which is one of the most established multi-view learning methods. Instead of modeling two views on the same set of data with different feature extractors or adversarial sample generation in conventional co-training \cite{blum1998combining,balcan2005co,qiao2018deep}, our two cross-domain views are explicitly provided by the segmentors with the correlated and complementary UDA and SSL tasks. Specifically, we construct the pseudo label of the unlabeled target sample based on the pixel-wise confident predictions of the other segmentor. Then, the segmentors are trained on the pseudo labeled data iteratively with an exponential MixUp decay (EMD) scheme for smooth propagation. Finally, the target segmentor carries out the target domain segmentation. 

The contributions of this work can be summarized as follows:

\noindent$\bullet$ We present a novel SSDA segmentation framework to exploit the different supervisions with the correlated and complementary asymmetric UDA and SSL sub-tasks, following a divide-and-conquer strategy. The knowledge is then integrated with confidence-aware pseudo-label based co-training.

\noindent$\bullet$ An EMD scheme is further proposed to mitigate the noisy pseudo label in early epochs of training for smooth propagation.

%A novel consensus-driven co-training scheme is proposed to integrate the multi-view supervisions with pseudo-label in a confidence-aware manner. %We show that ACT satisfies the $\epsilon$-expandability requirement of co-training.

\noindent$\bullet$ To our knowledge, this is the first attempt at investigating SSDA for medical image segmentation. Comprehensive evaluations on cross-modality brain tumor (i.e., T2-weighted MRI to T1-weighted/T1ce/FLAIR MRI) segmentation tasks using the BraTS18 database demonstrate superiority performance over conventional source-relaxed/source-based UDA methods.

\begin{figure}[t]
\begin{center}
\includegraphics[width=1\linewidth]{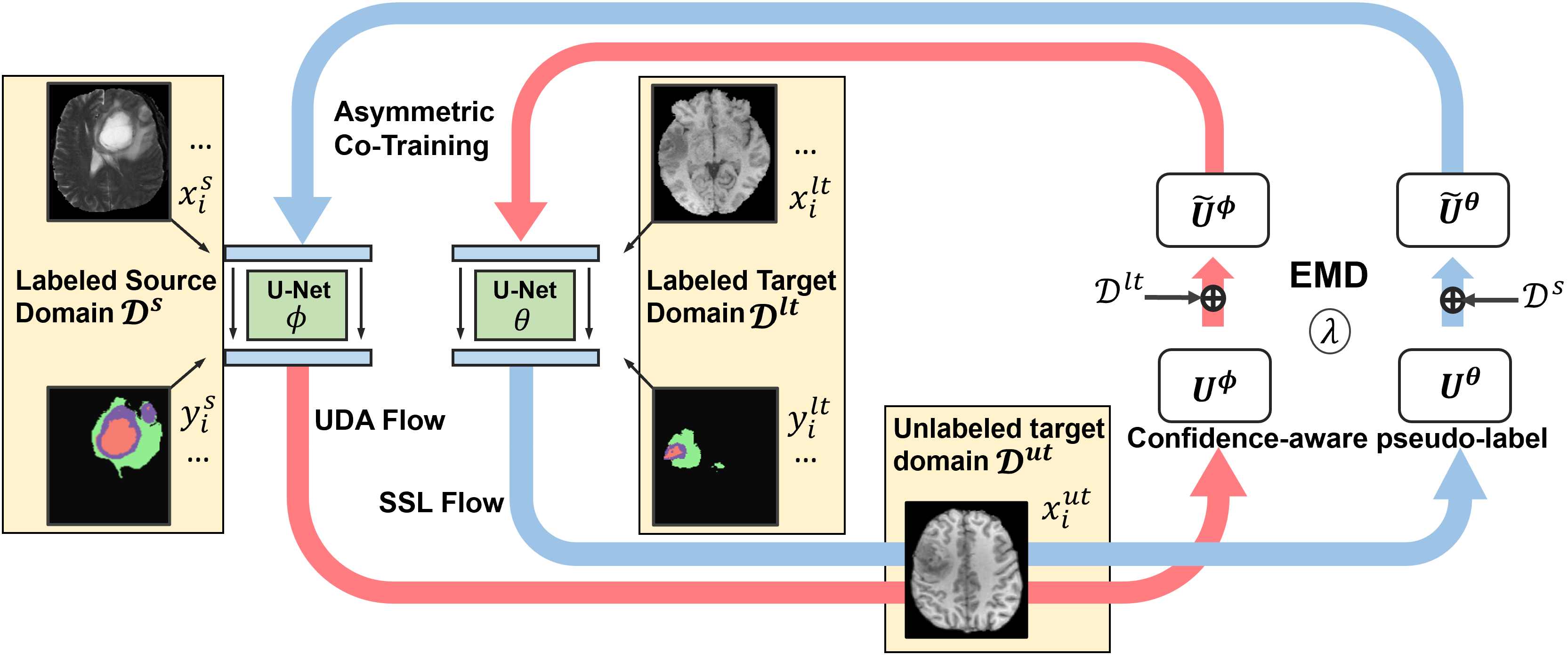}
\end{center}  
\caption{Illustration of our proposed ACT framework for SSDA cross-modality (e.g., T2-weighted to T1-weighted MRI) image segmentation. Note that only target domain specific segmentor $\theta$ will be used in testing.}  
\label{fig1}\end{figure}

\section{Methodology}

In our SSDA setting for segmentation, we are given a labeled source set $\mathcal{D}^s = \{(x^s_i,y^{s}_i)\}_{i=1}^{N^s}$, a labeled target set $\mathcal{D}^{lt} = \{(x^{lt}_i,y^{lt}_i)\}_{i=1}^{N^{lt}}$, and an unlabeled target set $\mathcal{D}^{ut} = \{(x^{ut}_i)\}_{i=1}^{N^{ut}}$, where ${N^s}$, ${N^{lt}}$, and ${N^{ut}}$ are the number of samples for each set, respectively. Note that the slice $x^s_i, x^{lt}_i$, and $x^{ut}_i$, and the segmentation mask labels $y_i^{s}$, and $y_i^{lt}$ have the same spatial size of $H\times W$. In addition, for each pixel $y^{s}_{i:n}$ or $y^{lt}_{i:n}$ indexed by $n\in\mathbb{R}^{H\times W}$, the label has $C$ classes, i.e., $y^{s}_{i:n}, y^{lt}_{i:n}\in\{1,\cdots,C\}$. There is a distribution divergence between source domain samples, $\mathcal{D}^s$, and target domain samples, $\mathcal{D}^{lt}$ and $\mathcal{D}^{ut}$. Usually, ${N^{lt}}$ is much smaller than ${N^s}$. The learning objective is to perform well in the target domain.

%Below, we introduce our Asymmetric Co-training (ACT) scheme in Subsec. 2.1, which can be further improved with a consensus-driven co-training scheme in Subsec. 2.2. The constraint for efficient ACT training is discussed in Subsec. 2.3. 

\subsection{Asymmetric Co-training for SSDA segmentation}

To decouple SSDA via a divide-and-conquer strategy, we integrate $\mathcal{D}^{ut}$ with either $\mathcal{D}^s$ or $\mathcal{D}^{lt}$ to form the correlated and complementary sub-tasks of UDA and SSL. We configure a cross-domain UDA segmentor $\phi$ and a target domain SSL segmentor $\theta$, which share the same objective of achieving a decent segmentation performance in $\mathcal{D}^{ut}$. The knowledge learned from the two segmentors is then integrated with ACT. The overall framework of this work is shown in Fig. \ref{fig1}. 
 
Conventional co-training has focused on two independent views of the source and target data or generated artificial multi-views with adversarial examples, which learns two classifiers for each of the views and teaches each other on the unlabeled data \cite{blum1998combining,qiao2018deep}. By contrast, in SSDA, without multiple views of the data, we propose to leverage the distinct yet correlated supervision, based on the inherent discrepancy of the labeled source and target data. We note that the sub-tasks and datasets adopted are different for the UDA and SSL branches. Therefore, all of the data subsets can be exploited, following well-established UDA and SSL solutions without interfering with each other. 

To achieve co-training, we adopt a simple deep pseudo labeling method \cite{wei2020theoretical}, which assigns the pixel-wise pseudo label $\hat{y}_{i:n}$ for $x^{ut}_{i:n}$. Though UDA and SSL can be achieved by different advanced algorithms, deep pseudo labeling can be applied to either UDA \cite{zou2019confidence} or SSL \cite{wei2020theoretical}. Therefore, we can apply the same algorithm to the two sub-tasks, thereby greatly simplifying our overall framework. We note that while a few methods \cite{xia2021uncertainty} can be applied to either SSL or UDA like pseudo labeling, they have not been jointly adopted in the context of SSDA.

Specifically, we assign the pseudo label for each pixel $x^{ut}_{i:n}$ in $\mathcal{D}^{ut}$ with the prediction of either $\phi$ or $\theta$, therefore constructing the pseudo labeled sets $U^{\phi}$ and $U^{\theta}$ for the training of another segmentor $\theta$ and $\phi$, respectively:  
\begin{align}
   U^{\phi} = \{(x^{ut}_{i:n},\hat{y}^{\phi}_{i:n}= \argmax_c p(c|x^{ut}_{i:n};\phi)); \text{ if } \max_c p(c|x^{ut}_{i:n};\phi) > \epsilon\},\\
   U^{\theta} = \{(x^{ut}_{i:n},\hat{y}^{\theta}_{i:n}= \argmax_c p(c|x^{ut}_{i:n};\theta)); \text{ if } \max_c p(c|x^{ut}_{i:n};\theta) > \epsilon\},
\end{align}
where $p(c|x^{ut}_{i:n};\theta)$ and $p(c|x^{ut}_{i:n};\phi)$ are the predicted probability of class $c\in\{1,\cdots,C\}$ w.r.t. $x^{ut}_{i:n}$ using $\theta$ and $\phi$, respectively. $\epsilon$ is a confidence threshold. Note that the low softmax prediction probability indicates the low confidence for training \cite{zou2019confidence,liu2021generative}. Then, the pixels in the selected pseudo label sets are merged with the labeled data to construct $\{\mathcal{D}^{s},U^{\theta}\}$ and $\{\mathcal{D}^{lt},U^{\phi}\}$ for the training of $\phi$ and $\theta$ with a conventional supervised segmentation loss, respectively. Therefore, the two segmentors with asymmetrical tasks act as teacher and student of each other to distillate the knowledge with highly confident predictions.

%We rank the softmax prediction probability for each class, and use adaptive threshold $\epsilon^\phi$ or $\epsilon^\theta$ to filter out 50\% unconfident predictions as \cite{liu2021generative,Zou_2019_ICCV}.    
 
%\subsection{Consensus-driven co-training} 
 
%DMT: Dynamic Mutual Training for Semi-Supervised Learning

%$\mathcal{D}^s = \{(x^s_i,y^{s}_i)\}_{i=1}^{N^s}$, a labeled target set $\mathcal{D}^{lt} = \{(x^{lt}_i,y^{lt}_i)\}_{i=1}^{N^{lt}}$, and an unlabeled target set $\mathcal{D}^{ut} = \{(x^{ut}_i)\}_{i=1}^{N^{ut}}$,

\begin{center}
\begin{algorithm}[t]
\SetAlgoLined
\scriptsize 
\caption{An iteration of the ACT algorithm.}
\SetKwInOut{Input}{Input}
\SetKwInOut{Output}{Output}
\textbf{Input:} batch size $N$, $\lambda$, $\eta$, $\epsilon$, $\mathcal{D}^s$, $\mathcal{D}^{lt}$, $\mathcal{D}^{ut}$, current network parameters $\omega_{\phi}$, $\omega_{\theta}$;\\
\textbf{{Sample}} $\{(x^s_i,y^{s}_i)\}_{i=1}^{N},\{(x^{lt}_i,y^{lt}_i)\}_{i=1}^{N},$ and$\{(x^{ut}_i)\}_{i=1}^{N}$from $\mathcal{D}^s,\mathcal{D}^{lt},$ and$\mathcal{D}^{ut},$ respectively;\\
\textbf{Initialize} $U^{\phi}=\emptyset$, $U^{\theta}=\emptyset$;\\
\For{$i\leftarrow 1$ \KwTo $N$}{
$\hat{y}^{\phi}_{i:n}= \argmax_c p(c|x^{ut}_{i:n};\phi);$ and $\hat{y}^{\theta}_{i:n}= \argmax_c p(c|x^{ut}_{i:n};\theta)$\\
\textbf{if} $\max_c p(c|x^{ut}_{i:n};\phi) > \epsilon$: \textbf{update} $U^{\phi} \leftarrow U^{\phi} \cup \{(x^{ut}_{i:n},\hat{y}^{\theta}_{i:n})$ with Eq. (1);\\
\textbf{if} $\max_c p(c|x^{ut}_{i:n};\theta) > \epsilon$: \textbf{update} $U^{\theta} \leftarrow U^{\theta} \cup \{(x^{ut}_{i:n},\hat{y}^{\theta}_{i:n})$ with Eq. (2);\\
}
\textbf{Obtain} $\tilde{U}^{\phi}=\{\text{EMD}({U}^{\phi}_i, \{(x^{lt}_i,y^{lt}_i)\}_{i=1}^{N}; \lambda)\}_{i=1}^{|{U}^{\phi}|\times N}$ with Eq. (3);\\
\textbf{Obtain} $\tilde{U}^{\theta}=\{\text{EMD}({U}^{\theta}_i, \{(x^{s}_i,y^{s}_i)\}_{i=1}^{N}; \lambda)\}_{i=1}^{|{U}^{\theta}|\times N}$ with Eq. (4);\\
\textbf{Update} $\omega_{\phi}\leftarrow \omega_{\phi}-\eta\nabla(\mathcal{L}(\omega_{\phi},\mathcal{D}^s)+\mathcal{L}(\omega_{\phi},\tilde{U}^{\theta}))$; $\omega_{\theta}\leftarrow \omega_{\theta}-\eta\nabla(\mathcal{L}(\omega_{\theta},\mathcal{D}^{lt})+\mathcal{L}(\omega_{\theta},\tilde{U}^{\phi}))$;\\
\Output{Updated network parameters $\omega_{\phi}$ and $\omega_{\theta}$.} 
\end{algorithm}\label{alg_code}
\end{center}

%Rank $\hat{y}^{\phi}_{i:n}$ or $\hat{y}^{\theta}_{i:n}$ for each class, and choose 50\% threshold as $\epsilon^\phi$ or $\epsilon^\theta$ for each class; \\

\subsection{Pseudo-label with Exponential MixUp Decay}  

Initially generated pseudo labels with the two segmentors are typically noisy, which is significantly acute in the initial epochs, thus leading to a deviated solution with propagated errors. Numerous conventional co-training methods relied on simple assumptions that there is no domain shift, and the predictions of the teacher model can be reliable and be simply used as ground truth. Due to the domain shift, however, the prediction of $\phi$ in the target domain could be noisy and lead to an aleatoric uncertainty \cite{der2009aleatory,kendall2017uncertainties,hu2019supervised}. In addition, insufficient labeled target domain data can lead to an epistemic uncertainty related to the model parameters \cite{der2009aleatory,kendall2017uncertainties,hu2019supervised}. 

To smoothly exploit the pseudo labels, we propose to adjust the contribution of the supervision signals from both labels and pseudo labels as the training progresses. Previously, vanilla MixUp \cite{zhang2017mixup} was developed for efficient data augmentation, by combining both samples and their labels to generate new data for training. We note that the MixUp used in SSL \cite{berthelot2019mixmatch,chen2021semi} adopted a constant sampling, and did not take the decay scheme for gradual co-training. Thus, we propose to gradually exploit the pseudo label by mixing up $\mathcal{D}^s$ or $\mathcal{D}^{lt}$ with pseudo labeled $\mathcal{D}^{ut}$, and adjust their ratio with the EMD scheme. For the selected ${U}^{\phi}$ and ${U}^{\theta}$ with the number of slices $|{U}^{\phi}|$ and $|{U}^{\theta}|$, we mix up each pseudo labeled image with all images from $\mathcal{D}^s$ or $\mathcal{D}^{lt}$ to form the mixed pseudo labeled sets $\tilde{U}^{\theta}$ and $\tilde{U}^{\phi}$. Specifically, our EMD can be formulated as:
\begin{align}
\tilde{U}^{\phi} =\{(\tilde{x}^{lt}_{i:n} = \lambda x^{lt}_{i:n} + (1-\lambda) x^{ut}_{i:n},  \tilde{y}^{lt}_{i:n} =\lambda y^{lt}_{i:n} + (1-\lambda) ~~\hat{y}^{\theta}_{i:n})\}_i^{|{U}^{\theta}|\times N},\\
\tilde{U}^{\theta} =\{(\tilde{x}^{s}_{i:n} = \lambda x^{s}_{i:n} + (1-\lambda) x^{ut}_{i:n}, \tilde{y}^{s}_{i:n} =\lambda y^{s}_{i:n} + (1-\lambda) ~~\hat{y}^{\phi}_{i:n})\}_i^{|{U}^{\phi}|\times N},
\end{align} where $\lambda =\lambda^0\text{exp}(-I)$ is the MixUp parameter with the exponential decay w.r.t. iteration $I$. $\lambda^0$ is the initial weight of ground truth samples and labels, which is empirically set to 1. Therefore, along with the increase over iteration $I$, we have smaller $\lambda$, which adjusts the contribution of the ground truth label to be large at the start of the training, while utilizing the pseudo labels at the later training epochs. Therefore, $\tilde{U}^{\phi}$ and $\tilde{U}^{\theta}$ gradually represent the pseudo label sets of ${U}^{\phi}$ and ${U}^{\theta}$. We note that the mixup operates on the image level, which is indicated by $i$. The number of generated mixed samples depends on the scale of ${U}^{\phi}$ and ${U}^{\theta}$ in each iteration and batch size $N$. With the labeled $\mathcal{D}^s$, $\mathcal{D}^{lt}$, as well as the pseudo labeled sets with EMD $\tilde{U}^{\phi}$ and $\tilde{U}^{\phi}$, we update the parameters of the segmentors $\phi$ and $\theta$, i.e., $\omega_{\phi}$ and $\omega_{\theta}$ with SGD as:
\begin{align}
\omega_{\phi}\leftarrow \omega_{\phi}-\eta\nabla(\mathcal{L}(\omega_{\phi},\mathcal{D}^s)+\mathcal{L}(\omega_{\phi},\tilde{U}^{\theta})),\\
\omega_{\theta}\leftarrow \omega_{\theta}-\eta\nabla(\mathcal{L}(\omega_{\theta},\mathcal{D}^{lt})+\mathcal{L}(\omega_{\theta},\tilde{U}^{\phi})),
\end{align} where $\eta$ indicates the learning rate, and $\mathcal{L}(\omega_{\phi},\mathcal{D}^s)$ denotes the learning loss on $\mathcal{D}^s$ with the current segmentor $\phi$ parameterized by $\omega_{\phi}$. The training procedure is detailed in Algorithm 1. After training, only the target domain specific SSL segmentor $\theta$ is used for testing.

%We mixup each image pair between $\mathcal{D}^s$ and ${U}^{\phi}$, or $\mathcal{D}^t$ and ${U}^{\theta}$. 

%\subsection{Constraint for efficient ACT training} 

\begin{figure}[t!]
\begin{center}
\includegraphics[width=1\linewidth]{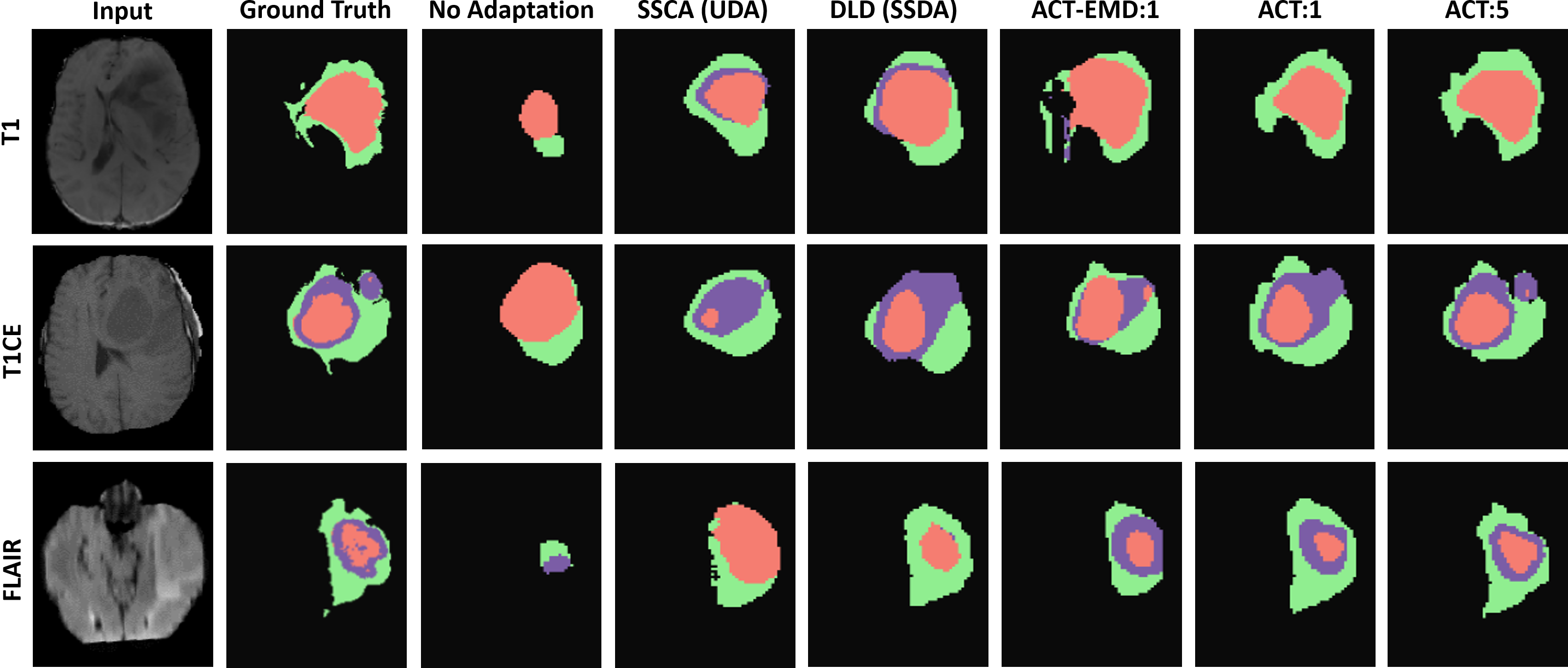}
\end{center}
\caption{Comparisons with other UDA/SSDA methods and ablation studies for the cross-modality tumor segmentation. We show target test slices of T1, T1ce, and FLAIR MRI from three subjects.}
\label{exp2}
\end{figure} 

%We implemented our method using PyTorch and tested it on an NVIDIA V100 GPU.

\section{Experiments and Results}

To demonstrate the effectiveness of our proposed SSDA method, we evaluated our method on T2-weighted MRI to T1-weighted/T1ce/FLAIR MRI brain tumor segmentation using the BraTS2018 database \cite{menze2014multimodal}.  We denote our proposed method as ACT, and used ACT-EMD for an ablation study of an EMD-based pseudo label exploration. %The implementation details are provided in \textbf{supplementary materials}. 

Of note, the BraTS2018 database contains a total of 285 patients \cite{menze2014multimodal} with the MRI scannings, including T1-weighted (T1), T1-contrast enhanced (T1ce), T2-weighted (T2), and T2 Fluid Attenuated Inversion Recovery (FLAIR) MRI. For the segmentation labels, each pixel belongs to one of four classes, i.e., enhancing tumor (EnhT), peritumoral edema (ED), necrotic and non-enhancing tumor core (CoreT), and background. In addition, the whole tumor covers CoreT, EnhT, and ED. We follow the conventional cross-modality UDA (i.e., T2-weighted to T1-weighted/T1ce/FLAIR) evaluation protocols \cite{zou2020unsupervised,han2022deep,liu2022self} for 8/2 splitting for training/testing, and extend it to our SSDA task, by accessing the labels of 1-5 target domain subjects at the adaptation training stage. All of the data were used in a subject-independent and unpaired manner. We used SSDA:1 or SSDA:5 to denote that one or five target domain subjects are labeled in training.

%Since typical clinical delineation of the brain tumors is carried out on T2-weighted MRI, we usually use T2-weighted MRI as our labeled source domain \cite{zou2020unsupervised,han2022deep,liu2022self}.

For a fair comparison, we used the same segmentor backbone as in DSA \cite{han2022deep} and SSCA \cite{liu2022self}, which is based on Deeplab-ResNet50. Without loss of generality, we simply adopted the cross-entropy loss as $\mathcal{L}$, and set the learning rate $\eta=1\mathrm{e}{-3}$ and confidence threshold $\epsilon=0.5$. Both $\phi$ and $\theta$ have the same network structure. For the evaluation metrics, we adopted the widely used DSC (the higher, the better) and Hausdorff distance (HD: the lower, the better) as in \cite{han2022deep,liu2022self}. The standard deviation was reported over five runs.

\begin{table}[t]
\centering
\caption{Whole tumor segmentation performance of the cross-modality UDA and SSDA. The supervised joint training can be regarded as an ``upper bound".} 
\resizebox{1\linewidth}{!}{
\begin{tabular}{l|c|cccc|cccc}
\hline

&  & DICE & Score &(DSC) & [$\%$] $\uparrow$  & Hausdorff & Distance & (HD) & [mm] $\downarrow$ \\ \cline{2-10}

{Method}& Task  & \textbf{T1}  & \textbf{FLAIR} & \textbf{T1CE}& \textbf{Ave}  &  \textbf{T1}  & \textbf{FLAIR} & \textbf{T1CE} & \textbf{Ave}  \\ \hline \hline

Source Only &No DA &4.2& 65.2& 6.3 &27.7$\pm$1.2& 55.7 &28.0& 49.8 &39.6$\pm$0.5\\\hline 
Target Only &SSL:5 &43.8& 54.6& 47.5 &48.6$\pm$1.7& 31.9 &29.6& 35.4 &32.3$\pm$0.8\\\hline\hline 
 
SIFA \cite{chen2019synergistic}&UDA &51.7& 68.0& 58.2 &59.3$\pm$0.6 &19.6& 16.9& 15.0& 17.1$\pm$0.4\\
DSFN \cite{zou2020unsupervised}&UDA &57.3& 78.9 &62.2& 66.1$\pm$0.8& 17.5& 13.8 &15.5 &15.6$\pm$0.3\\
DSA \cite{han2022deep}&UDA &57.7 &81.8 &62.0& 67.2$\pm$0.7 &14.2 &8.6 &13.7& 12.2$\pm$0.4\\
SSCA \cite{liu2022self}&UDA &59.3 &82.9  &  63.5& 68.6$\pm$0.6 & 12.5& 7.9& 11.2& 11.5$\pm$0.3  \\ \hline \hline

SLA \cite{wang2020alleviating}&SSAD:1  &64.7& 82.3 &66.1& 71.0$\pm$0.5 &12.2& 7.1& 10.5& 9.9$\pm$0.3\\

DLD \cite{chen2021semi}&SSAD:1  &65.8& 81.5 &66.5& 71.3$\pm$0.6 &12.0& 7.1& 10.3& 9.8$\pm$0.2\\\hline

ACT&SSAD:1  & \textbf{69.7} &\textbf{84.5}& \textbf{69.7}& \textbf{74.6$\pm$0.3}& \textbf{10.5} &\textbf{5.8}& \textbf{10.0}& \textbf{8.8$\pm$0.1}\\

ACT-EMD&SSAD:1  & 67.4 &83.9 &69.0& 73.4$\pm$0.6 &10.9 &6.4 &10.3 &9.2$\pm$0.2\\\hline

ACT&SSAD:5   & \textbf{71.3} &\textbf{85.0}& \textbf{70.8}& \textbf{75.7$\pm$0.5}& \textbf{10.0} &\textbf{5.2}& \textbf{9.8}& \textbf{8.3$\pm$0.1}\\

ACT-EMD&SSAD:5  & 70.3 &84.4& 69.8& 74.8$\pm$0.4 &10.4 &5.7 &10.2 &8.8$\pm$0.2\\\hline \hline

Joint Training&Supervised & 73.2 &85.6& 72.6& 77.1$\pm$0.5& 9.5& 4.6 &9.2& 7.7$\pm$0.2\\\hline

\end{tabular}}\label{tab1}  
\end{table}

\begin{table}[t]
\centering
\caption{Detailed comparison of Core/EnhT/ED segmentation. Results are averaged over three tasks including T2-weighted to T1-weighted, T1CE, and FLAIR MRI with the backbone as in \cite{han2022deep,liu2022self}.} 
\resizebox{1\linewidth}{!}{
\begin{tabular}{l|c|ccc|ccc}
\hline

& & DICE & Score~~(DSC) & [$\%$] $\uparrow$  & Hausdorff & Distance (HD) & [mm] $\downarrow$ \\ \cline{2-8}

{Method}& Task&\textbf{CoreT} & \textbf{EnhT} & \textbf{ED}  & \textbf{CoreT} & \textbf{EnhT} & \textbf{ED}  \\ \hline \hline

Source Only &No DA &20.6$\pm$1.0 &39.5$\pm$0.8 &41.3$\pm$0.9  &54.7$\pm$0.4 &55.2$\pm$0.6 &42.5$\pm$0.4 \\\hline

Target Only &SSL:5&27.3$\pm$1.1 &38.0$\pm$1.0 &40.2.3$\pm$1.3  &51.8$\pm$0.7 &52.3$\pm$0.9 &46.4$\pm$0.6 \\\hline \hline
    
DSA \cite{han2022deep}  & UDA  &57.8$\pm$0.6& 44.0$\pm$0.6 &56.8$\pm$0.5 &25.8$\pm$0.4 &34.2$\pm$0.3& 25.6$\pm$0.5\\
SSCA \cite{liu2022self}  & UDA &  58.2$\pm$0.4&  44.5$\pm$0.5& 60.7$\pm$0.4& 26.4$\pm$0.2& 32.8$\pm$0.2&23.4$\pm$0.3  \\ \hline \hline   

SLA \cite{wang2020alleviating}  & SSDA:1&58.9$\pm$0.6&48.1$\pm$0.5 &65.4$\pm$0.4&24.5$\pm$0.1 &  27.6$\pm$0.3 &20.3$\pm$0.2 \\
DLD \cite{chen2021semi} & SSDA:1&60.3$\pm$0.6 &48.2$\pm$0.5 &66.0$\pm$0.3& 24.2$\pm$0.2 &  27.8$\pm$0.1&19.7$\pm$0.2   \\ \hline

ACT  & SSDA:1&\textbf{64.5$\pm$0.3} &\textbf{52.7$\pm$0.4} &\textbf{69.8$\pm$0.6}& \textbf{20.0$\pm$0.2}&  \textbf{24.6$\pm$0.1} &\textbf{16.2$\pm$0.2} \\\hline

ACT & SSDA:5&\textbf{{66.9}$\pm$0.3} &\textbf{{54.0}$\pm$0.3}  &\textbf{{71.2}$\pm$0.5} &\textbf{{18.4}$\pm$0.4}  & \textbf{{23.7}$\pm$0.2} &\textbf{{15.1}$\pm$0.2} \\\hline\hline

Joint Training & Supervised  & 70.4$\pm$0.3& 62.5$\pm$0.2& 75.1$\pm$0.4& 15.8$\pm$0.2 &22.7$\pm$0.1& 13.0$\pm$0.2\\\hline  

\end{tabular}}\label{tab2}
\end{table}

The quantitative evaluation results of the whole tumor segmentation are provided in Table~\ref{tab1}. We can see that SSDA largely improved the performance over the compared UDA methods \cite{han2022deep,liu2022self}. For the T2-weighted to T1-weighted MRI transfer task, we were able to achieve more than 10\% improvements over \cite{han2022deep,liu2022self} with only one labeled target sample. Recent SSDA methods for natural image segmentation \cite{wang2020alleviating,chen2021semi} did not take the balance between the two labeled supervisions into consideration, easily resulting in a source domain-biased solution in case of limited labeled target domain data, and thus did not perform well on target domain data \cite{saito2019semi}. In addition, the depth estimation in \cite{hoyer2021improving} cannot be applied to the MRI data. Thus, we reimplemented the aforementioned methods~\cite{wang2020alleviating,chen2021semi} with the same backbone for comparisons, which is also the first attempt at the medical image segmentation. Our ACT outperformed \cite{wang2020alleviating,chen2021semi} by a DSC of 3.3\% w.r.t. the averaged whole tumor segmentation in SSDA:1 task. The better performance of ACT over ACT-EMD demonstrated the effectiveness of our EMD scheme for smooth adaptation with pseudo-label. We note that we did not manage to outperform the supervised joint training, which accesses all of the target domain labels, which can be considered an ``upper bound" of UDA and SSDA. Therefore, it is encouraging that our ACT can approach joint training with five labeled target subjects. In addition, the performance was stable for the setting of $\lambda$ from 1 to 10. 

In Table~\ref{tab2}, we provide the detailed comparisons for more fine-grained segmentation w.r.t. CoreT, EnhT, and ED. The improvements were consistent with the whole tumor segmentation. The qualitative results of three target modalities in Fig.~\ref{exp2} show the superior performance of our framework, compared with the comparison methods.

\begin{figure}[t]
\begin{center}
\includegraphics[width=1\linewidth]{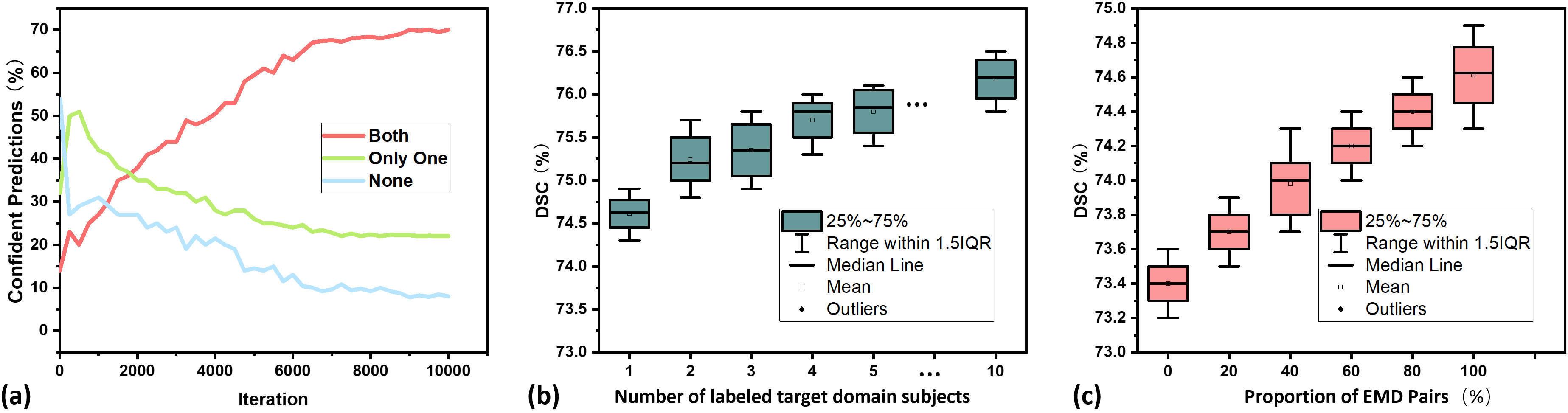}
\end{center}
\caption{Analysis of our ACT-based SSDA on the whole tumor segmentation task. (a) The proportion of testing pixels that both, only one, or none of the segmentors have high confidence on (b) the performance improvements with a different number of labeled target domain training subjects, and (c) a sensitivity study of changing different proportion of EMD pairs of $|\tilde{U}^{\phi}|\times N$ and $|\tilde{U}^{\theta}|\times N$.} 
\label{exp3}
\end{figure}

In Fig. \ref{exp3}(a), we analyzed the testing pixel proportion change along with the training that has both, only one, and none of two segmentor pseudo-labels, i.e., the maximum confidence is larger than $\epsilon$ as in Eq.~(1). We can see that the consensus of the two segmentors keeps increasing, by teaching each other in the co-training scheme for knowledge integration.  ``Both" low rates, in the beginning, indicate $\phi$ and $\theta$ may provide a different view based on their asymmetric tasks, which can be complementary to each other. The sensitivity studies of using a different number of labeled target domain subjects are shown in Fig.~\ref{exp3}(b). Our ACT was able to effectively use $\mathcal{D}^{lt}$. In Fig. \ref{exp3}(c), we show that using more EMD pairs improves the performance consistently.

\section{Conclusion}

This work proposed a novel and practical SSDA framework for the segmentation task, which has the great potential to improve a target domain generalization with a manageable labeling effort in clinical practice. To achieve our goal, we resorted to a divide-and-conquer strategy with two asymmetric sub-tasks to balance between the supervisions from source and target domain labeled samples. An EMD scheme is further developed to exploit the pseudo-label smoothly in SSDA. Our experimental results on the cross-modality SSDA task using the BraTS18 database demonstrated that the proposed method surpassed the state-of-the-art UDA and SSDA methods. %In addition, our method can be performed with supervised joint training on limited labeled datasets.

\section*{Acknowledgements}

This work is supported by NIH R01DC018511, R01DE027989, and P41EB022544.

\bibliographystyle{splncs04}
\bibliography{egbib}

\end{document}